\documentclass[11pt,a4paper]{article}
\usepackage{array}
\usepackage[hyperref]{naaclhlt2018}
\usepackage{times}
\usepackage{latexsym}
\usepackage{graphicx}
\usepackage{url}
\usepackage{verbatim}
\usepackage{amsmath}
\usepackage{multicol}
\usepackage{multirow}
\usepackage{float}

\newcolumntype{L}[1]{>{\raggedright\let\newline\\\arraybackslash\hspace{0pt}}m{#1}}
\newcolumntype{C}[1]{>{\centering\let\newline\\\arraybackslash\hspace{0pt}}m{#1}}
\newcolumntype{R}[1]{>{\raggedleft\let\newline\\\arraybackslash\hspace{0pt}}m{#1}}

\aclfinalcopy 

\title{Syntactic Patterns Improve Information Extraction for Medical Search \vspace{-2.0em}}
\vspace{-2.5em}
\author{Roma Patel$^{12}$, Yinfei Yang\thanks{* now at Google Inc.}, Iain Marshall$^3$,
{\bf Ani Nenkova$^1$~} and {\bf Byron C. Wallace$^4$}\\
 $^1${Department of Computer and Information Science, University of Pennsylvania}\\
 $^2${Department of Computer Science, Rutgers University-Camden}\\
      $^3${Department of Primary Care and Public Health Sciences, King’s College London}\\
      $^4${College of Computer and Information Science, Northeastern University }\\
 }
\date{}
\begin{document}
\maketitle

\begin{abstract}
\vspace{-.5em} 
Medical professionals search the published literature by specifying the type of {\em patients}, the medical {\em intervention(s)} and the {\em outcome} measure(s) of interest. In this paper we demonstrate how features encoding syntactic patterns improve the performance of state-of-the-art sequence tagging models (both linear and neural) for information extraction of these medically relevant categories. We present an analysis of the type of patterns exploited, and the semantic space induced for these, i.e., the distributed representations learned for identified multi-token patterns. We show that these learned representations differ substantially from those of the constituent unigrams, suggesting that the patterns capture contextual information that is otherwise lost. 
\end{abstract}

\section{Introduction}
\vspace{-.5em}

The efficacy of medical treatments depends on patient characteristics, treatment administration details (e.g., dosage) and the measures or outcomes used to quantify treatment success. These criteria should be precisely defined when searching the medical literature ~\cite{richardson1995well, heneghan2013evidence,miller2001enhancing}. Unfortunately, these aspects are not usually described in a structured way.
Abstracts with explicit category headings ~\cite{nakayama2005adoption} partially address this, but these are not standardized nor uniform. Automated solutions are thus emerging to better support medical search, including methods for: identifying sentences containing key pieces of clinical information \cite{DBLP:journals/jmlr/WallaceKSZM16}; summarization ~\cite{sarker2016query}; identifying contradictory claims in medical articles ~\cite{alamri2016corpus}; and information retrieval system prototypes that harness this type of information \cite{DBLP:conf/naacl/BoudinND10,DBLP:conf/ecir/BoudinSN10}.

Several studies have assessed the use of the PICO framework \cite{huang2006evaluation, demner2007answering}. Our task is also to identify spans of text describing \emph{PICO elements} 
i.e., the participants ({\sc p}), interventions ({\sc i})/comparators ({\sc c}), and outcomes ({\sc o}) in the abstracts of articles reporting findings from randomized controlled trails (RCTs). We exploit the availability of structured abstracts in the medical domain: from these coarse (multi-)sentence labels we derive patterns typically used in bootstrap methods for entity recognition and relation extraction \cite{DBLP:conf/aaai/CarlsonBKSHM10}. We incorporate these patterns into supervised sequence labeling models to improve the identification of {\sc p}, {\sc i} and {\sc o} spans in new texts. 
Below we show examples of each extraction type: patterns are bolded and target PICO description spans italicized.
The extracted patterns disambiguate fairly well the type of information expressed in the segment when individual words (e.g., ``children"), do not. 
\vspace{.1em}
\noindent
\textbf{(P)} The trial included \emph{230} \textbf{children with} \emph{Stage-IV lymphoblastic leukemia} ...

\vspace{.1em}
\noindent
\textbf{(I)}  In Group I, the \textbf{children were treated} with \emph{prednisone} ... 

\vspace{.1em}
\noindent
\textbf{(O)} .. reported that Group 2 \textbf{children underwent} fewer \emph{isolated bone marrow relapses} ..
\vspace{.1em}

We explore three strategies for exploiting extracted patterns in a state-of-the-art LSTM-CRF sequence tagging model ~\cite{lample2016neural, ma-hovy:2016:P16-1}: as additional features at the CRF layer; as one-hot indicators concatenated to distributed representations of words; and as individual units embedded in a semantic space shared with words. The second representation improves recall for two extraction tasks, and the third improves precision for all three tasks. 
We analyze the induced semantic space to show that patterns capture contextual information that is otherwise lost.
 
\section{Data}
For training sequence tagging models we use a corpus of 4,741 medical article abstracts with manual crowd-sourced annotations for {\sc p}, {\sc i}, and {\sc o} sequences. For testing we use a set of 191 abstracts annotated for {\sc p}, {\sc i}, and {\sc o} by medical professionals. There are 18,849 (831), 44,329 (1,808), 41,454 (1,711) variable length sequences for {\sc p}, {\sc i}, and {\sc o} in the training (testing) data.\footnote{The complete details of the corpus, along with inter-annotator analysis and links for download of the full corpus will be described in a forthcoming paper and eventually made available here: \url{http://www.byronwallace.com/EBM_abstracts_data}.}

For minimally supervised extraction of $n$-gram patterns, we use structured abstracts in which the authors describe different aspects of their work under targeted headings. We retrieved the headings and associated sections automatically from abstracts in {\sc xml} format (downloaded from PubMed\footnote{\url{https://www.nlm.nih.gov/databases/download/pubmed_medline.html}}). In general abstracts are structured idiosyncratically (often as Introduction, Methods, Results, Discussion). We capitalized on the minority of abstracts that used the explicit Participants, Intervention and Outcome headings. We obtained 50,000 segments for each of these three categories.

\section{Patterns extraction and analysis}
\vspace{-.5em}
We extract syntactic patterns associated with each of the extraction types using AutoSlog-TS ~\cite{Riloff:1996:AGE:1864519.1864542}, which consumes two sets of text: one relevant to an extraction domain and one irrelevant. In our case the relevant sets are the 50K {\sc p}, {\sc i}, and {\sc o} sections, respectively, from the structured abstracts described above. The irrelevant set is a mix of 25K of the other two categories.

AutoSlog-TS generates $n$-gram patterns from input texts that capture the context of all noun phrases appearing as subject, direct and indirect object, or in a prepositional phrase. Each of these patterns is scored with the estimated probability that it occurred in an instance from the relevant set (out of all occurrences of the pattern), scaled by the number of times the pattern occurs \cite{riloff2004introduction}. Common patterns that tend to occur in relevant sentences thus receive relatively high scores. We filter out patterns that contain digits, and those that occur fewer than 10 times in the structured abstract texts. Of the remaining patterns, we preserve those with probability 0.8 or higher of occurring with the relevant class. This yields 3,499, 3,898 and 2,386 patterns associated with {\sc p}, {\sc i} and {\sc o}, respectively.

The vast majority of patterns are bigrams: 90\% for {\sc p}, 81\% for {\sc i} and 86\% {\sc o}. Fewer than 0.5\% of the $n$-grams for each type are trigrams, and the remaining are unigrams. Examples of extracted patterns include: \textit{women\_who}, \textit{years\_of} and \textit{diagnosed\_with} for {\sc p}; \textit{patients\_received} and \textit{performed\_after} for {\sc i}; and \textit{scale\_of}, \textit{patients\_reported} and \textit{rate\_of} for {\sc o}. 

The majority (82.86\%) of the extracted $n$-gram patterns comprise a combination of a content word and a function/stopword token.\footnote{We use stopwords from CoreNLP \cite{DBLP:conf/acl/ManningSBFBM14}.} For example, the patterns \textit{patients\_with, patients\_who} or \textit{patients\_from} are associated with the condition that a patient had, while \textit{patients\_were, patients\_in} or \textit{patients\_received} describe the treatment they received. Function words provide disambiguating context for otherwise ambiguous words; this aids text classification and information retrieval ~\cite{riloff1995little}, and here we use them to improve sequence tagging models. 
 
\begin{table*}[h]
\small
\begin{center}
\begin{tabular}{c c c c | c c c | c c c}
&  \multicolumn{3}{c|}{\textbf{Precision} } &  \multicolumn{3}{c|}{\textbf{Recall} } &  \multicolumn{3}{c}{\textbf{F1} } \\
\cline{2-10}
 \bf Model & \bf P & \bf I & \bf O & \bf P & \bf I & \bf O & \bf P & \bf I & \bf O \\ \hline
CRF & 70.29 & 47.01 & 64.78 & 38.75 & 43.89 & 10.47 & 49.95 & 45.39 &  18.02\\
CRF-Pattern & \bf 73.3 & \bf 52.1 & \bf 66.37 & \bf 40.62 & \bf 45.41 & \bf 44.07 & \bf 52.27 & \bf 48.52 & \bf 52.96 \\
\hline
LSTM-CRF & 62.27 & 52.37 & \bf 47.91 & 49.48 & 40.49 & \bf 36.16 & 55.14 & 45.67 & \bf 41.21 \\
LSTM-CRF-Pattern (best) & \bf 76.10 & \bf 58.25 & 44.66 & \bf 64.75 & \bf 43.39 &  35.20 & \bf 69.97 & \bf 49.74 & 39.69 \\
\hline
Before CRF & 61.87 & 38.65 & 46.27 & 41.45 & 23.8 & 37.27 & 49.64 & 29.55 & 41.28 \\
Before BiLSTM & \bf 76.10 & \bf 58.25 & 44.66 & \bf 64.75 & 43.39 &  35.20 & \bf 69.97 & \bf 49.74 & 39.69 \\
Embedding & 55.18 & 51.07 & 44.30& 54.24 & \bf 47.41 & \bf 41.60 & 54.71 &  49.17 & \bf 42.91 \\
\hline
\end{tabular}
\end{center}
\vspace{-.75em}
\caption{Models for extracting {\bf P}articipants, {\bf I}ntervention and {\bf O}utcomes with and without pattern features, evaluated via token-level precision, recall and F1 scores. The first and second groups of rows report results for CRF and LSTM-CRF models without and with pattern features. The bottom group reports results achieved using different means of incorporating pattern features in neural models.}
\label{tab:results}
\vspace{-.75em}
\end{table*}

\section{Patterns + linear CRF}
\vspace{-.5em}
For supervised IE models, we first consider including $n$-gram patterns as features in a linear-chain CRF ~\cite{lafferty2001conditional}. The standard set of token-level features used in the model include word identity, POS tag (from CoreNLP), and a list of binary features indicating whether the token is a 
digit, title (i.e., the first token only is uppercase), uppercase word, hyphenated word, or if the token is a punctuation mark (colon, fullstop or another symbol). In addition, features for the current token include the identity of the previous and next words, and the immediately preceeding and following bi- and trigrams.

For the pattern-augmented CRF (\textbf{CRF-Pattern}), we add nine binary features that indicate if the current token and the immediately preceeding/following bigrams are one of the AutoSlog-TS patterns associated with a given extraction type.\footnote{We ignore trigram patterns as they constitute $<$$0.5$\% of identified patterns.} There are three indicators, for {\sc p}, {\sc i} and {\sc o} respectively. For the context bigrams, a feature is 1 if the bigram is one of the bigram patterns associated with this extraction type, 0 otherwise. The remaining three indicators have value 1 if the current token is one of the unigram patterns associated with a given type. For example, the nine features for the token ``chronic'' in the sequence \textit{patients with chronic sinus issues} will be [1,0,0|0,0,0|0,0,0] because \textit{patients\_with} is one of the bigrams associated with the {\sc p} type, the word "chronic" does not match any of the unigram patterns and ``sinus issues'' does not match any of the bigram patterns. Table \ref{tab:results} (top) reports the performance on the test data of the original CRF model, and the one augmented with pattern features. Including patterns yields consistent and considerable improvements in both precision and recall. 

\section{Patterns + LSTM-CRF}
\vspace{-.5em}

LSTM-CRF models \cite{lample2016neural,ma-hovy:2016:P16-1} for sequence tagging are general in that they do not require feature engineering. Instead, the features representing each token in the CRF are generated by a bi-directional LSTM. To generate this representation the LSTM consumes distributed word representations as input and outputs vector representations describing words {\em in context} (the bi-LSTM runs one LSTM in each direction, concatenating outputs). This vector is passed to a CRF layer for prediction. Character-level information for each word is incorporated by running a bi-LSTM over the characters of each word ~\cite{lample2016neural}. We used the IO tagging scheme. We set the hidden state dimensions to 200 and dropout to 0.5. We did not perform gradient clipping. We used the Adam optimizer \cite{kingma2014adam} with learning rate = 0.001. 

We consider three alternatives for extending this model with patterns. The first two use the indicator features describing the presence of patterns in the context, similar to those we described above for the linear CRF model. The difference is where these features are introduced: immediately before the CRF layer, concatenated with the output of the LSTMs (\textbf{Before CRF}), or as part of the input to the LSTM, concatenated to the distributed word and character representations (\textbf{Before LSTM}).  
We use ~\newcite{moen2013distributional}'s release of 200 dimensional word vectors trained over 5.5 billion words from medical articles as pre-trained word embeddings as input to the LSTM. We use the same set of hyperparamaters for the LSTM as used in \newcite{lample2016neural}, and do not optimize these for the present extraction tasks. The third alternative (\textbf {Embedding}) treats the patterns as collocations; we derive embedded representations for them as a unit, the way collocations are treated in \newcite{mikolov2013distributed}. In training and during prediction each occurrence of a pattern in the input is treated as a single token with a corresponding distributed representation. Character-level representations are concatenated to word representations and the output of the LSTM cells is passed to the CRF to make predictions (as above). 

For these embeddings, we collected 6 million PubMed abstracts ($\sim$1.4 billion words) filtering for only Human RCTs and used this to train word vectors using the Word2Vec tool \cite{mikolov2013efficient}, inducing 200-dimensional vectors using the Skip-Gram model, where our vocabulary now consists of the learned $n$-gram patterns as single units, along with other unigrams. We then test these embedding representations by using them as input to our neural model for the structured prediction task.

\begin{table*}[h]
\small
\begin{center}
\begin{tabular}{c c c}
\bf $n$-gram & \multicolumn{1}{c}{\bf similar to $n$-gram} & \multicolumn{1}{c}{\bf similar to unigram}  \\ \hline
have\_\textbf{children} & 1: \textit{marry} 2: \textit{conceive} 3: \textit{breast-feed} & 1: \textit{adults} 2: \textit{adolescents} 3: \textit{toddlers}  \\
 & 4: \textit{be\_pregnant} 5: \textit{have\_surgery} & 4: \textit{youngsters} 5: \textit{school-age} \\
  
\textbf{condition}\_at & 1: \textit{status\_at} 2: \textit{features\_at}  3: \textit{outcome\_at}   & 1: \textit{circumstance}  2: \textit{conditions} 3: \textit{malady}  \\
 &  4: \textit{qol\_at} 5: \textit{outcomes}  & 4: \textit{ailment}  5: \textit{situation} \\
 
  \textbf{filled}\_with & 1: \textit{covered\_with} 2: \textit{mixed\_with}  3: \textit{sealed\_with}   & 1: \textit{sealed}  2: \textit{obturated} 3: \textit{enclosed}  \\
 &  4: \textit{suspended} 5: \textit{immersed\_in}  & 4: \textit{enclosing}  5: \textit{fill} \\
 
 side\_\textbf{effects} & 1: \textit{toxicities} 2: \textit{side-effect}  3: \textit{complications}   & 1: \textit{effect}  2: \textit{Effects} 3: \textit{action}  \\
 &  4: \textit{AEs} 5: \textit{nausea}  & 4: \textit{impact}  5: \textit{influence} \\
\end{tabular}
\end{center}
\vspace{-.75em}
\caption{Example illustrating the shift in semantic space realized using pattern embeddings. For each of the listed $n$-grams, we report the top 5 most similar words to: (1) the $n$-gram pattern embedding, and, (2) the most relevant constituent $n$-gram i.e., the word in bold font.}
\label{tab:similarities}
\vspace{-1em}
\end{table*}

\section{Discussion of results}
\vspace{-.5em}

Table \ref{tab:results} reports the performance of the LSTM-CRF model achieved using each of the three strategies for incorporating pattern features discussed above. Inserting the pattern indicator features before the CRF layer yields the worst performance. Compared to the generic LSTM-CRF model, its $F$-measure is lower or the same for all three extraction categories, {\sc p}, {\sc i}, {\sc o}. 

Including the pattern features as input to the LSTM or as part of the embedding leads to substantial improvements over the baseline model, and this despite the smaller dataset over which pattern embeddings were learned: compared to the LSTM-CRF without pattern features, the former markedly improves precision for {\sc p} and {\sc i}, while the latter improves the recall for all three types. In terms of $F$-measure, best results for {\sc p} and {\sc i} are achieved by inserting the pattern features as input to the LSTM, with about 15\% and 4\% absolute improvement. For {\sc o}, the best $F$-measure is achieved by incorporating patterns as part of the embeddings, yielding 1\% absolute improvement.

The linear CRF and its variant enriched with pattern feature has the best precision, outperforming the LSTM-CRF models, but worse recall. It may still be useful for scenarios in which high precision extraction is needed.

 \section{Semantics of pattern embeddings}
 \vspace{-.5em}
We established that syntactic patterns can markedly improve the extraction of patient, intervention and outcome descriptions in medical abstracts. We now turn to an analysis of how the patterns fit into the semantic space of word embeddings. Our goal is to quantify the extent to which including pattern representations changes which words will be considered similar to the pattern, but not to the words that compose it.

To this end, we find the ten words most similar (under cosine similarity) to each pattern, and those most similar to the individual words these comprise, in the embedding space. We analyze the size of the intersection of these two sets for all patterns ($\sim$10,000). To simplify the comparison we consider only the constituent word that has the largest intersection of similar words with the pattern of interest. The size of the intersection theoretically ranges from 0 to 10, but on average there is only one word overlap between the words most similar with the pattern and those most similar with the constituent word. For the majority (61\%) of the pattern--constituent word pairs, there is no overlap between the top 10 most similar words. To make this discussion more concrete, Table \ref{tab:similarities} provides examples of the top 5 most similar words to select bigram patterns and the constituent unigram with greatest overlap, shown in italics. The patterns encode disambiguating context that was previously lost in unigram representations.  

\begin{figure}
\includegraphics[width=0.5\textwidth]{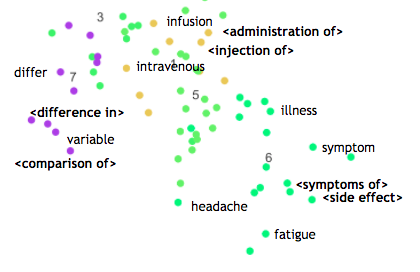}
\vspace{-1.5em}
\caption{Scatter of PCA-reduced embeddings clustered using K-means. $<>$ brackets show the syntactic pattern n-grams given by Autoslog-TS that are embedding in the same space as unigrams.}
\vspace{-0.5em}
\label{figure:pattern-embeddings}
\end{figure}

\begin{table*}
\small
\begin{center}
\begin{tabular}{c c c c | c c c | c c c}
&  \multicolumn{3}{c|}{\textbf{Precision} } &  \multicolumn{3}{c|}{\textbf{Recall} } &  \multicolumn{3}{c}{\textbf{F1} } \\
\cline{2-10}
 \bf Model & \bf P & \bf I & \bf O & \bf P & \bf I & \bf O & \bf P & \bf I & \bf O \\ 
\hline
LSTM-CRF & 62.27 & 52.37 & \bf 47.91 & 49.48 & 40.49 & 36.16 & 55.14 & 45.67 & \bf 41.21 \\
LSTM-CRF (Bigrams) & 64.41 & 53.37 &  43.20 & 50.33 & 41.24 &  \bf 37.32 & 59.91 & 46.52 & 40.04 \\
LSTM-CRF (Autoslog) & \bf 76.10 & \bf 58.25 & 44.66 & \bf 64.75 & \bf 43.39 &  35.20 & \bf 69.97 & \bf 49.74 & 39.69 \\
\hline
\end{tabular}
\end{center}
\caption{Results to illustrate syntactic nature of Autoslog bigrams. Row 1 shows results of baseline model with no added features. Row 2 shows results of the model that uses all bigrams as features and Row 3 shows results of the model that uses only Autoslog extracted bigrams as features. Features are added before the LSTM, as incorporated in the best working model from Table 1.}
\label{tab:results-bigram}
\end{table*}

Finally, we present a scatter of learned embeddings, reduced via IncrementalPCA\footnote{We use the implementation in \emph{scikit-learn} \cite{scikit-learn}.} in Figure \ref{figure:pattern-embeddings}. Embedded patterns cluster more intuitively than their content words alone. For example, the patterns \textit{injection\_of} and \textit{administration\_of} cluster together, along with other topically similar unigrams such as \textit{infusion} and \textit{intravenous} that may all correspond to Intervention terms. Similarly, \textit{side\_effect} is very different from its constituent words \textit{side} or \textit{effect}, and moreover, clusters with actual side effects like \textit{headache} and \textit{fatigue} that patients may suffer from in the course of a trial. 

\section{Syntactic patterns vs bigrams}
Our experiments show that using these bigram features extracted by AutoSlog improves model predictions. AutoSlog takes a fundamentally syntax-driven approach to identifying patterns, which suggests the discovered patterns (and associated performance boost) is due to exploiting syntax. However, the performance gains could also be due to additional contextual information that bigrams and larger $n$-grams provide over unigrams alone, rather than their syntactic properties. 

We therefore performed an experiment to assess the influence of the syntactic AutoSlog bigrams, as compared to general bigram features. We consider the same data used as input to AutoSlog, i.e., 50,000 segments for the three categories {\sc p}, {\sc i}, and {\sc o}. In the same setup, we decompose sentences within each category into bigrams, and collect bigram counts in the respective categories. We calculate precision for each category by collapsing the other two categories, similar to the AutoSlog procedure. We use the same threshold values as AutoSlog for filtering, i.e., we remove bigrams that occur fewer than 10 times or that have a score $<$0.8 of occuring with the target class out of all occurrences. This procedure for identifying predictive bigrams yields a notably larger number of bigrams (30k) than AutoSlog ($\sim$10K). Table \ref{tab:results-bigram} shows that while using generic bigrams as features sometimes leads to small improvements, the AutoSlog induced pattern bigrams result in substantially better performance. This suggests that the exploitation of syntactic structure in identifying patterns is indeed important. We also compare the performance of word2vec embeddings for unigrams and bigrams, and extended with collocations and syntactic patterns, trained on exactly the same data. In the experiments reported in Table \ref{tab:results}, the unigram embeddings are trained on a larger dataset of generic medical text while the patterns are trained on a smaller set of medical abstracts describing RCTs. In addition here we compare the AutoSlog patterns with collocations discovered by word2vec. Representing collocations leads to markedly lower F-score (Table \ref{tab:sameTrainingData}). Representing bigrams leads to prediction performance better than that with collocations, but worse than unigrams.

Standard unigram representations that we trained work better than the off-the-shelf medical representations, possibly because they were trained specifically on abstracts of papers reporting the conduct and results of RCTs and thus better fit the abstracts we are analyzing. Most importantly, the LSTM-CRF with syntactic pattern embeddings results in the best observed performance.

\begin{table}[h]
\small
\begin{center}
\begin{tabular}{c|c|c|c|c}
\small
\bf Embedding & \bf Vocabulary & \bf P & \bf I & \bf O \\ \hline
\noalign{\vskip 1mm} 
Unigram & 947,670 & 54.31 & 46.19 & \textbf{42.68} \\
Bigram & 9,326,144 & 52.01 & 43.71 & 38.77 \\
collocation & 1,254,863 & 50.31 & 40.21 & 40.21 \\
Pattern & 949,112 & \textbf{54.71} & \textbf{47.68} & 42.27 \\
\hline
\end{tabular}
\end{center}
\caption{LSTM-CRF predictions on word embeddings trained on the same 6 million documents. Column 1 shows the type of embedding, column 2 shows the size of the vocabulary and columns 3-5 show F1 score.}
\vspace{-1em}
\label{tab:sameTrainingData}
\end{table}

\section{Conclusions}
\vspace{-.5em}
We presented a method for exploiting abundant unlabeled biomedical texts to generate minimally supervised extraction patterns that improve generic supervised models for sequence tagging in this domain. We explored alternative ways to incorporating the patterns in both linear and neural tagging models. In the latter, we analyzed the changes in semantic space that likely explain the observed gains in predictive performance.

\section{Acknowledgements}
\vspace{-.5em}
This work was supported in part by the National Cancer Institute (NCI) of the National
Institutes of Health (NIH), award number UH2CA203711 and the National Science Foundation (NSF), award number CCF-1433220. 
\bibliography{naaclhlt2018}
\bibliographystyle{acl_natbib}
\end{document}